\documentclass{article}

\usepackage{arxiv}

\usepackage[utf8]{inputenc} 
\usepackage[T1]{fontenc}    
\usepackage{hyperref}       
\usepackage{url}            
\usepackage{booktabs}       
\usepackage{amsfonts}       
\usepackage{nicefrac}       
\usepackage{microtype}      
\usepackage{lipsum}		
\usepackage{graphicx}
\usepackage{float}
\usepackage{natbib}
\usepackage{doi}
\usepackage{amsmath,amssymb}
\usepackage{multirow}
\usepackage{makecell}
\usepackage{xcolor}
\usepackage{colortbl}
\usepackage{tabularx}
\usepackage{adjustbox}
\usepackage{placeins}
\usepackage{xpatch}

\makeatletter
\renewcommand{\@maketitle}{%
  \vbox{%
    \hsize\textwidth
    \linewidth\hsize
    \vskip 0.1in
    \@toptitlebar
    \centering
    {\LARGE\scshape\@title\par}%
    \@bottomtitlebar
    \begin{tabular}[t]{c}%
      \rule{\z@}{10\p@}\@author
    \end{tabular}%
    \vskip 0.12in
  }%
}
\makeatother

\makeatletter
\ifdefined\XeTeXversion
  \newcommand*\PDRVLA@linkoff{}
  \newcommand*\PDRVLA@linkon{}
\else\ifdefined\pdfrunninglinkoff
  \let\PDRVLA@linkoff\pdfrunninglinkoff
  \let\PDRVLA@linkon\pdfrunninglinkon
\else\ifdefined\luatexversion
  \newcommand*\PDRVLA@linkoff{\pdfextension linkstate 1\relax}
  \newcommand*\PDRVLA@linkon{\pdfextension linkstate 0\relax}
\else
  \PackageError{PDRVLA}{This PDF engine cannot suspend page-spanning links}
    {Use XeLaTeX, or pdfLaTeX/LuaLaTeX from TeX Live 2021 or newer.}%
\fi\fi\fi
\AtBeginDocument{%
  \ifdefined\pdfannot@link@off@@
    \let\pdfannot@link@off@@\PDRVLA@linkoff
    \let\pdfannot@link@on@@\PDRVLA@linkon
  \else
    \xpatchcmd{\@outputpage}
      {\vfil\color@hbox}
      {\vfil\PDRVLA@linkoff\color@hbox}
      {}{\PackageError{PDRVLA}{Header link suspension patch failed}{Check the output routine.}}%
    \xpatchcmd{\@outputpage}
      {{\@thehead}\color@endbox}
      {{\@thehead}\color@endbox\PDRVLA@linkon}
      {}{\PackageError{PDRVLA}{Header link resumption patch failed}{Check the output routine.}}%
    \xpatchcmd{\@outputpage}
      {\footskip\color@hbox\normalcolor}
      {\footskip\color@hbox\normalcolor\PDRVLA@linkoff}
      {}{\PackageError{PDRVLA}{Footer link suspension patch failed}{Check the output routine.}}%
    \xpatchcmd{\@outputpage}
      {{\@thefoot}\color@endbox}
      {{\@thefoot}\PDRVLA@linkon\color@endbox}
      {}{\PackageError{PDRVLA}{Footer link resumption patch failed}{Check the output routine.}}%
  \fi
}
\makeatother

\newcommand{\reporttablestyle}{\small\setlength{\tabcolsep}{4pt}\renewcommand{\arraystretch}{1.12}}
\usepackage[ruled,vlined,linesnumbered]{algorithm2e}

\newcommand{\ourmodel}{PDRVLA}

\newcommand{\NA}{--}

\definecolor{ourrow}{gray}{0.92}

\title{PDRVLA: A Vision--Language--Action Model with Physical Dynamics Representations}

\author{
    \normalfont Hanqing Wang\textsuperscript{2,*} \quad
    Zhaotian Cai\textsuperscript{2,*} \quad
    Ruichuan An\textsuperscript{1,*} \quad
    Bohan Zeng\textsuperscript{1,*} \quad
    Tao Zhang\textsuperscript{2,*} \quad
    Yilin Chen\textsuperscript{2,*} \\[0.3em]
    \normalfont Xu Jiang\textsuperscript{2} \quad
    Kaiquan Chen\textsuperscript{2} \quad
    Chongyang Wu\textsuperscript{2} \quad
    Haoran Li\textsuperscript{2} \quad
    Fan Yang\textsuperscript{2} \quad
    Mingan Lin\textsuperscript{2} \\[0.3em]
    \normalfont Wentao Zhang\textsuperscript{1,\ensuremath{\dagger}} \quad
    Zenan Zhou\textsuperscript{2,\ensuremath{\dagger}} \quad
    Youzheng Wu\textsuperscript{2} \quad
    Xiaodong He\textsuperscript{2} \\[0.6em]
    {\normalfont \textsuperscript{1}Peking University \quad
    \textsuperscript{2}JD.com}
}

\date{}

\renewcommand{\headeright}{}
\renewcommand{\undertitle}{}
\renewcommand{\shorttitle}{PDRVLA}
\hypersetup{
pdftitle={PDRVLA: A Vision-Language-Action Model with Physical Dynamics Representations},
pdfsubject={Robot learning and vision-language-action models},
pdfauthor={Hanqing Wang, Zhaotian Cai, Ruichuan An, Bohan Zeng, Tao Zhang, Yilin Chen, Xu Jiang, Kaiquan Chen, Chongyang Wu, Haoran Li, Fan Yang, Mingan Lin, Wentao Zhang, Zenan Zhou, Youzheng Wu},
pdfkeywords={PDRVLA, vision-language-action, physical dynamics, future prediction, robot learning},
}

\begin{document}
\raggedbottom
\maketitle
\begingroup
\renewcommand{\thefootnote}{\fnsymbol{footnote}}
\footnotetext[1]{Equal contribution.}
\footnotetext[2]{Corresponding authors: \href{mailto:zhouzenan@jd.com}{\nolinkurl{zhouzenan@jd.com}}, \href{mailto:wentao.zhang@pku.edu.cn}{\nolinkurl{wentao.zhang@pku.edu.cn}}.}
\endgroup

\vspace{10pt}
\begin{abstract}
Vision--Language--Action (VLA) models generate robot actions from visual observations and language instructions, but their understanding of physical dynamics remains limited. Video world models can capture scene dynamics, yet generating future videos introduces high latency into the control loop. We introduce \ourmodel{}, whose lightweight Physical Dynamics Module extracts physical tokens from vision--language features to condition action generation without online video generation. The module learns through direct prediction of future RGB, depth, and optical-flow features, with auxiliary supervision through a frozen pretrained video world model. Training relies entirely on publicly available data, expanded through human-to-robot synthesis, subtask annotation, and embodied chain-of-thought (ECoT) annotation. Across four simulation benchmarks, \ourmodel{} achieves success rates of 99.3\% on LIBERO, 89.9\% on LIBERO-Plus, 93.6\% on RoboTwin~2.0, and 81.5\% on SimplerEnv. Ablations show improved future prediction with world-model supervision and evaluate how subtask conditioning, future object-pose supervision, and ECoT training affect action success.
\end{abstract}

\vspace{10pt}
\begingroup
\setlength{\intextsep}{6pt}
\setlength{\abovecaptionskip}{5pt}
\begin{figure}[!ht]
    \centering
    \includegraphics[trim=0 30bp 0 10bp,clip,width=\linewidth]{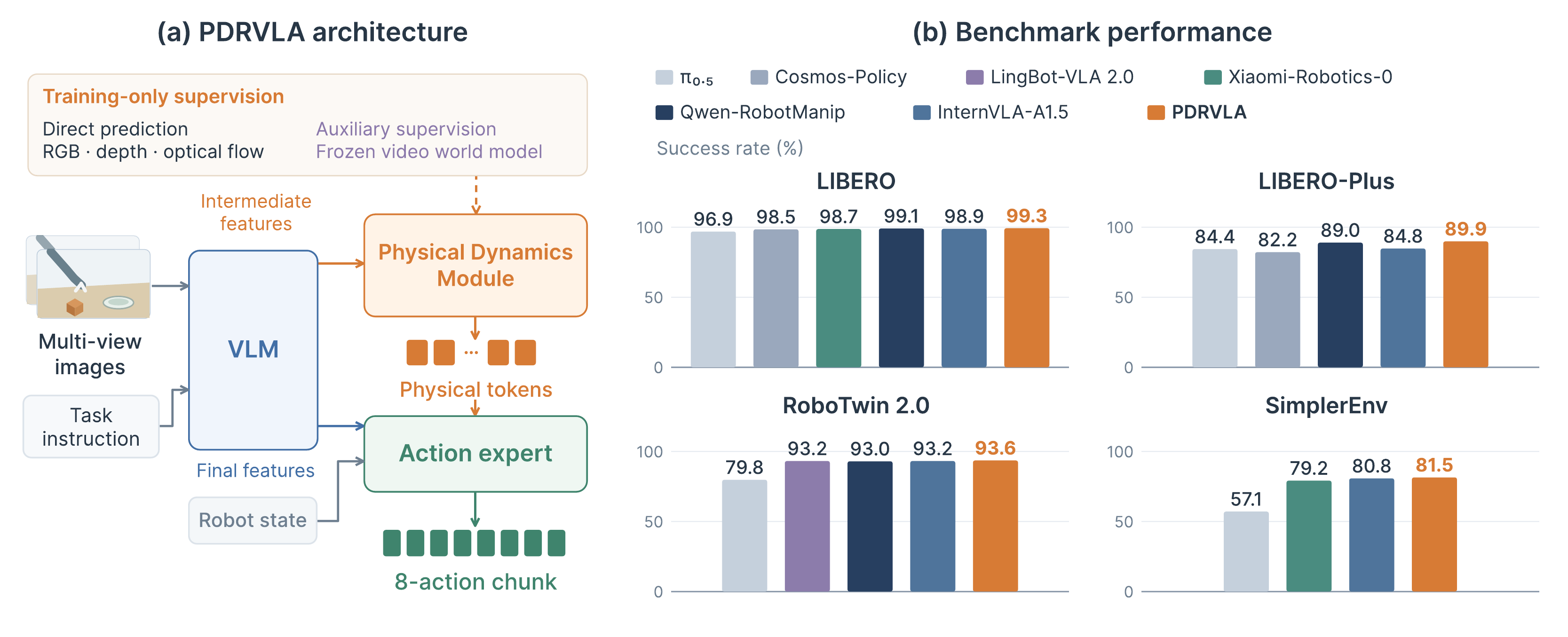}
    \caption{\textbf{PDRVLA overview.} (a) Simplified architecture. (b) Success rates on four simulation benchmarks.}
    \label{fig:intro_overview}
\end{figure}
\endgroup


\clubpenalties 3 10000 10000 0\relax
\section{Introduction}

Vision--Language--Action (VLA) models connect visual observations and language instructions to robot control, drawing on pretrained vision--language models (VLMs) and demonstrations from heterogeneous robots~\citep{black2024pi0,intelligence2025pi05}. Their understanding of the physical world, however, remains limited. For example, a recent analysis reports weak encoding of object physical properties across several VLA policies and unreliable adaptation of their motions to changes in object mass~\citep{xu2026physicalunderstanding}. For manipulation, recognizing the current scene is only part of the problem: grasping, moving, and placing objects also change their configurations and spatial relationships. Anticipating how a scene evolves during interaction can provide useful context for action selection, motivating representations that connect current observations with future appearance, geometry, and motion.
\clubpenalties 0\relax

Pretrained video world models offer a complementary way to learn physical dynamics by predicting how scenes change over time~\citep{wan2025}. Incorporating these models into robot policies can connect future prediction with action generation~\citep{kim2026cosmospolicy}. However, generating future videos within the control loop adds latency to action generation. Recent approaches instead use video prediction during training to shape compact representations that can condition a policy without online video generation~\citep{cai2026internvlaa15,tang2026worldtokens}.

Following this direction, we introduce \ourmodel{}, a VLA model with a lightweight Physical Dynamics Module (Figure~\ref{fig:intro_overview}). The module extracts a fixed set of physical tokens from intermediate VLM features. The action expert uses these tokens alongside the final-layer VLM features to generate actions. We train the module with two complementary objectives. Direct supervision uses future RGB, depth, and optical-flow latent features as targets, encouraging the tokens to encode future appearance, geometry, and motion. A frozen pretrained video world model conditioned on the physical tokens provides auxiliary supervision for temporal prediction. Joint policy training then combines action and vision--language (VL) objectives with continued direct future prediction. The resulting policy uses physical tokens for action generation without running the video model during control.

Our training corpus is built entirely from publicly available datasets. It combines real-robot, handheld UMI, and simulated trajectories with general and robotics-oriented VL data, and is expanded through egocentric human-to-robot synthesis. Subtask and embodied chain-of-thought (ECoT) annotations provide additional supervision for task stages, scene understanding, progress assessment, and next motion. Together, these sources support training across heterogeneous embodiments and complementary forms of action and language supervision.

We evaluate \ourmodel{} on LIBERO~\citep{liu2023libero}, LIBERO-Plus~\citep{fei2025liberoplus}, RoboTwin 2.0~\citep{chen2025robotwin}, and SimplerEnv~\citep{li2025simpler}, achieving success rates of 99.3\%, 89.9\%, 93.6\%, and 81.5\%, respectively. Ablation studies show that world-model-assisted supervision improves future prediction. Additional experiments examine the effects of subtask conditioning, future object-pose supervision, and ECoT annotations on policy performance.

The report makes three contributions:
\begin{itemize}
    \item We introduce \ourmodel{}, whose lightweight Physical Dynamics Module extracts compact physical tokens from intermediate VLM features to complement the final-layer features used for action generation. We train these representations through direct prediction of future RGB, depth, and optical-flow features, with auxiliary supervision from a frozen video world model. Direct future prediction continues during joint policy training, while deployment uses the learned physical tokens without running the video generator.
    \item We assemble a cross-embodiment training corpus entirely from publicly available datasets, integrating real-robot, handheld UMI, and simulated demonstrations with general and robotics-oriented VL data. Our data pipeline standardizes heterogeneous trajectories, expands embodiment coverage through egocentric human-to-robot synthesis, and adds subtask and ECoT annotations, supporting joint learning of robot actions and language responses about task stages, scene context, progress, and next motion.
    \item We evaluate \ourmodel{} on four simulation benchmarks and conduct targeted analyses of physical-representation learning and policy performance. The results show improved future prediction with world-model assistance and characterize the effects of subtask conditioning, future object-pose supervision, and ECoT annotations on action success across tasks and inference settings.
\end{itemize}

\section{Data}
\label{sec:data}
\subsection{Corpus Overview}
Our training corpus combines VLA data with VL data. The VLA portion pairs observations and task instructions with control targets, grounding the model in robot behavior across diverse embodiments and environments. The VL portion extends perceptual, spatial, and task-level reasoning beyond the coverage of action trajectories.

We organize the corpus into five categories. The VLA portion comprises real-robot trajectories, UMI demonstrations, and simulated interactions, which provide complementary trade-offs in physical fidelity, collection cost, embodiment coverage, and scale. The VL portion comprises general image- and video-language data together with spatial and embodied examples more closely aligned with manipulation. Table~\ref{tab:data_overview} summarizes the composition and scale of each category; individual sources are described in Sections~\ref{sec:vla_data} and~\ref{sec:vl_data}.

\begin{table}[!htbp]
\centering
    \reporttablestyle

\caption{Core training corpus by category.}
\label{tab:data_overview}
\begin{tabular}{@{}ll@{}}
\toprule
\textbf{Category} & \textbf{Statistics} \\
\midrule
\multicolumn{2}{@{}l}{\textit{VLA Data}} \\
Real-Robot & 1.26M trajectories / 14{,}893 hours \\
\addlinespace
UMI & 1.22M trajectories / 10{,}259 hours \\
\addlinespace
Simulation & 1.87M trajectories / 8{,}868 hours \\
\midrule
\multicolumn{2}{@{}l}{\textit{VL Data}} \\
General VL & 8.05M VL examples \\
\addlinespace
Spatial \& Embodied VL & 23.63M VL examples \\
\bottomrule
\end{tabular}
\end{table}

In addition to aggregating existing datasets, we derive new training data through synthesis and annotation. Egocentric human-to-robot synthesis converts EgoDex demonstrations~\citep{hoque2026egodex} into visually and kinematically adapted trajectories for multiple robot embodiments. Subtask annotation assigns a stage label to each frame and produces subtask-prediction VL examples, while ECoT annotation produces keyframe-level VL examples describing the scene, task progress, and next motion. Table~\ref{tab:augmentation} summarizes these derived data.

\begin{table}[!htbp]
\centering
    \reporttablestyle
\caption{Statistics of synthesized and annotated training data.}
\label{tab:augmentation}
\begin{tabular}{@{}ll@{}}
\toprule
\textbf{Augmentation} & \textbf{Statistics} \\
\midrule
Egocentric Conversion & 0.90M robot trajectories / 2{,}646 hours \\
Subtask Annotation & 1.79B annotated frames / 1.1M VL examples \\
ECoT & 2.56M annotated keyframes \\
\bottomrule
\end{tabular}%
\end{table}

\subsection{VLA Data}
\label{sec:vla_data}

\subsubsection{Real-Robot Data}
\label{sec:real_robot_data}
Real-robot trajectories expose the model to real-world dynamics, sensor variation, and contact-rich interactions. We aggregate public datasets spanning different robots, environments, and task families.

Our single-arm tabletop and in-the-wild manipulation data come from RoboSet~\citep{bharadhwaj2024roboagent}, DROID~\citep{khazatsky2024droid}, BridgeData V2~\citep{walke2023bridgedata}, BC-Z~\citep{jang2022bc}, and RH20T~\citep{fang2023rh20t}. Together, they cover diverse scenes, objects, operators, and manipulation skills, including pick-and-place, articulated-object interaction, and tool use.

We extend the embodiment coverage with bimanual and multi-platform data from RoboMIND V1/V2~\citep{robomind2025,hou2025robomind}, RoboCOIN~\citep{wu2025robocoin}, and ABC-130K~\citep{abc2026}; humanoid, arm, and dexterous-hand data from Humanoid Everyday~\citep{zhao2025humanoid} and the Unitree UniFoLM releases~\citep{unitree2026unifolmwbt,unitree2026unifolmdex1diverse,unitree2026unifolmdex1,unitree2026unifolmbrainco,unitree2026unifolmz1,unitree2026unifolmdex3}; and large-scale multi-robot data from AgiBot World Colosseo~\citep{bu2025agibot,contributors2024agibotworldrepo,agibotworld2026}.

REASSEMBLE~\citep{sliwowski2025reassemble} contributes contact-rich assembly and disassembly trajectories, while FuSe~\citep{jones25fuse} and MolmoAct~\citep{molmoact2025} provide additional manipulation trajectories. 

Selected embodiment-specific subsets from the LeRobot community collections~\citep{hfvlavla2025communityv1,hfvlavla2025communityv2,hfvlavla2025communityv3} add further cross-institution and cross-embodiment coverage.

\subsubsection{UMI Data}
UMI data are collected with a handheld gripper and wrist-mounted camera rather than by teleoperating a particular robot~\citep{chi2024umi}. The resulting demonstrations record real-world interaction as camera observations and end-effector pose trajectories without tying the motion to a robot's joint configuration.

We use six UMI datasets with different scales, capture systems, and task coverage: UMI~\citep{chi2024umi}, FastUMI~\citep{wu2024fast}, FastUMI-100K~\citep{liu2025fastumi100k}, Hy-Embodied-0.5-VLA-Data~\citep{zhang2026hyembodied05vlavisionlanguageactionmodelsrealworld}, HiFi-UMI-2K~\citep{simpleai2026hifiumi}, and RealOmni-Open Data~\citep{genrobot2025realomni}. After source-specific filtering, they contribute approximately 1.22 million episodes and 10{,}259 hours.

Because UMI demonstrations do not provide robot joint-space commands, we use their temporally aligned end-effector poses directly to construct the unified Cartesian action representation in Section~\ref{sec:data_processing}.

\subsubsection{Simulation Data}
Simulation complements real-world data with controllable, reproducible interaction and access to ground-truth states and success signals. It also makes it practical to vary tasks and embodiments without additional hardware collection.

The simulation corpus covers three complementary settings: standardized skills and long-horizon tasks from ManiSkill2~\citep{maniskill22023}, Meta-World~\citep{metaworld2021}, and RoboCasa~\citep{robocasa2024}; large-scale cross-embodiment trajectories from InternData-A1~\citep{tian2026interndata}, GR00T-X-Embodiment-Sim~\citep{nvidia2025gr00txsim}, and InternVLA-M1~\citep{internvlam12025}; and bimanual or dexterous manipulation from RoboTwin 2.0~\citep{chen2025robotwin} and DexMimicGen~\citep{dexmimicgen2025}.

\subsection{VL Data}
\label{sec:vl_data}
\subsubsection{General VL Data}
The general VL mixture includes vision-centric instruction data from Cambrian-1~\citep{tong2024cambrian}; captioning and open-ended visual question answering from COCO-Caption~\citep{lmmslab2026cococaption} and VQAv2~\citep{huggingfacem42026vqav2}; and image- and video-reasoning examples from OneThinker~\citep{onethinker2026}.

\subsubsection{Spatial and Embodied VL Data}
Spatial and embodied VL data bring the supervision closer to manipulation. We divide them into spatial reasoning over general imagery and embodied or procedural reasoning drawn from robot data, simulation, and web video.

The spatial-reasoning group covers spatial question answering, geometric relations, coordinate-aware grounding, referring-expression localization, and multi-view or 3D reasoning. It includes SpaceThinker~\citep{remyxai2026spacethinker}, OpenSpaces~\citep{remyxai2026openspaces}, SpaceOm~\citep{remyxai2026spaceom}, RefSpatial~\citep{jingkunan2026refspatial}, data released with Scaling Spatial Intelligence with Multimodal Foundation Models~\citep{sensenova_si2025}, and cross-view correspondence examples linking egocentric, third-person, and robot-camera observations~\citep{crossviewpoint2026}.

The embodied and procedural group includes point supervision for object, free-space, and affordance localization from RoboPoint~\citep{yuan2024robopoint}; visual question answering and long-horizon reasoning over manipulation videos from Robo2VLM~\citep{robo2vlm2026} and RoboVQA~\citep{sermanet2024robovqa}; interleaved perception and reasoning from EO-1~\citep{qu2025eo1}; physical-commonsense supervision from Cosmos-Reason1~\citep{nvidia2025cosmosreason1}; and web-derived procedural supervision for step decomposition and progress reasoning from ProcCorpus-60M~\citep{feng2026procvlm}.

\subsection{Data Processing}
\label{sec:data_processing}

Trajectories from different datasets vary in action format, coordinate conventions, hardware names, and video configuration. We standardize action representations, filter invalid trajectories and frames, unify embodiment metadata, and align and encode visual observations. Processing rules account for differences in hardware and collection procedures.

\paragraph{Action Representation.}
To standardize end-effector (EEF) poses across datasets, we convert positions to meters and orientations to rotation matrices, and normalize the coordinate axes to a right-handed convention with $+x$ forward, $+y$ left, and $+z$ up. Each manipulator's poses remain relative to the arm base or the world frame, as specified by the source dataset. We express its current and target poses in the same reference frame and record that frame in the metadata.

We represent EEF actions as changes from the current pose to a target pose. The target is the dataset-provided absolute EEF target pose when available, or the next observed pose after resampling otherwise. For a given manipulator, let $r$ denote its reference frame, $e$ the EEF frame, and $t$ a timestep after resampling. The current and target positions are $\mathbf{p}_{r,t},\mathbf{p}_{r,t}^*\in\mathbb{R}^3$, and their orientations are $R_{re,t},R_{re,t}^*\in SO(3)$. The superscript $*$ denotes the target. Here, $R_{re,t}$ maps vectors from the EEF frame $e$ into reference frame $r$. The action consists of a translation delta and a rotation delta, both expressed in $r$:
\begin{equation}
    \Delta a^{\mathrm{eef}}_t=
    \left(\Delta\mathbf{p}_{r,t},\Delta R_{r,t}\right),\qquad
    \Delta\mathbf{p}_{r,t}=\mathbf{p}_{r,t}^*-\mathbf{p}_{r,t},\quad
    \Delta R_{r,t}=R_{re,t}^*R_{re,t}^{\mathsf T},
    \label{eq:pose-delta}
\end{equation}

For grippers with opening measurements, we normalize the current and target openings to $[0,1]$ using the gripper's physical travel range, with 0 denoting fully closed and 1 fully open. The gripper action is the normalized absolute target opening.

\paragraph{Trajectory Filtering.}
We reject entire episodes that are too short, lack usable action targets or state sequences, contain values whose units cannot be recovered, or have image and state streams that cannot be aligned. We also reject episodes whose task descriptions conflict with the source metadata. Filtering thresholds are set separately for each dataset or subset to account for differences in hardware and collection procedures.

For otherwise usable episodes, frames containing non-finite values, abnormal repeated observations, abrupt spikes, or out-of-range values are excluded from supervision but retained at their original temporal positions to preserve image--state--action alignment. If the corruption cannot be isolated to individual frames, we reject the entire episode.

\paragraph{Embodiment Metadata Standardization.}
Robot hardware names vary across datasets, and the same arm can be paired with different grippers or dexterous hands. We record robot, arm, and end-effector types separately in \texttt{robot\_type}, \texttt{arm\_type}, and \texttt{hand\_type}. We standardize aliases such as \texttt{panda} and \texttt{franka\_panda} to a common arm name while retaining the identity of the attached end effector. Missing or inconsistent metadata are checked against source documentation and recorded images or videos.

\paragraph{Visual Processing.}
Camera observations and robot states may be recorded at different rates. We align them at the lowest rate among the streams, capped at 30~Hz. At most four camera views are retained per episode, prioritizing wrist and third-person views, with view selection configured for each dataset. The selected videos are resized to 640 pixels on the longest side and encoded with a maximum group-of-pictures (GOP) size of 2 frames to accelerate random frame access. Post-processing checks verify that video frames remain aligned with robot states and actions.

\subsection{Data Synthesis and Annotation}
We expand the training corpus through human-to-robot synthesis and language annotation. The former adapts egocentric demonstrations into robot observations and actions; the latter adds subtask labels and ECoT descriptions of the scene, task progress, and next motion to existing trajectories.

\subsubsection{Egocentric Human-to-Robot Synthesis}
Human egocentric data cover diverse real-world scenes and manipulation tasks, but human hands and arms differ from robot manipulators in both appearance and kinematics. We therefore adapt both the recorded motion and the visual observations before using these demonstrations for VLA training.

We apply the human-to-robot synthesis pipeline of Qwen-RobotManip~\citep{qwen2026robotmanip} to EgoDex~\citep{hoque2026egodex}, generating demonstrations for 12 robot-arm models: Panda, UR5e, UR10e, FR3, Kinova Gen3, IIWA, ARX-L5, xArm7, ViperX, WidowX, Piper, and YAM.

\paragraph{Motion Retargeting.} We map EgoDex's per-frame 3D hand keypoints to an end-effector (EEF) pose trajectory and gripper commands. The EEF trajectory is mapped into each target robot's coordinate system, and inverse kinematics (IK) produces the corresponding arm-joint trajectory. We discard demonstrations with too few high-confidence hand keypoints and retargeted trajectories whose EEF poses, computed through forward kinematics, deviate excessively from the desired poses.

\paragraph{Visual Adaptation.} We segment the human hands and arms with SAM3~\citep{carion2026sam} and use ProPainter~\citep{zhou2023propainter} to reconstruct the background in the masked regions across the video. We render the target robot arm and gripper using the retargeted motion, then composite them into the reconstructed video. DA3 depth estimates~\citep{lin2025depth} determine which scene surfaces occlude the rendered robot during compositing. The resulting videos, arm-joint trajectories, and gripper commands form the synthetic VLA examples.

After filtering and standardization, the synthesized corpus contains 896{,}832 bimanual robot trajectories, totaling approximately 2{,}646 hours. These trajectories derive from 74{,}813 distinct EgoDex demonstrations, each of which may contribute variants for multiple target robots. Because the robot variants share a common source pool, the resulting datasets have similar sizes across robot models, with approximately 74{,}000--75{,}000 trajectories and 219--221 hours per target arm model.


\subsubsection{Subtask Annotation}
\label{sec:subtask-aug}
Long-horizon manipulation tasks contain stages with distinct intermediate goals. Following prior work on hierarchical task structure~\citep{zhang2024universal,chen2023sequential}, we segment trajectories into ordered subtasks and assign a subtask label to each frame. These annotations yield both subtask-annotated VLA trajectories and VL examples about task stages and transitions.

\paragraph{Subtask Segmentation.}
Source-provided subtask annotations are reused when available; the remaining trajectories are automatically annotated with Qwen3.8-27B~\citep{qwen38} on sparsely sampled observations. We sample one frame per second and increase the sampling rate for short trajectories to provide more observations. Observations at each sampled timestamp come from a selected camera view, with multiple views spatially concatenated when needed, such as two wrist-camera views for bimanual interactions.

The sampled observations are divided into sliding windows of at most 128 frames with a 16-frame overlap. For each window, the model predicts subtask descriptions and segment boundaries. For each sampled frame in an overlap, we use the prediction from the window whose midpoint is closest to that frame. We merge these predictions into an episode-level segmentation and propagate the segment labels to every frame, producing subtask-annotated VLA trajectories. Frames after the last predicted boundary receive a \texttt{done} label indicating the end of the annotated sequence.

\begin{figure}[!htbp]
    \centering
    \includegraphics[width=\linewidth]{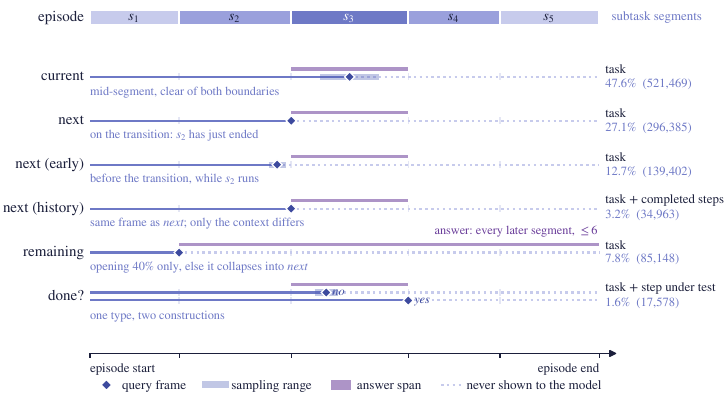}
    \caption{
    Six types of subtask VL examples, differing in query timing, answer content, and subtask-history context. No observations after the query frame are provided.
    }
    \label{fig:subtask-qtypes}
\end{figure}

\paragraph{Subtask VL Examples.}
Following $\pi_{0.5}$~\citep{intelligence2025pi05}, we further convert these annotations into VL examples that ask the model to identify the current subtask from a visual observation and the episode-level instruction. Additional queries ask for the next subtask, the remaining subtask sequence, or whether the current segment has reached its annotated end. Figure~\ref{fig:subtask-qtypes} summarizes six variants that differ in query time and in whether prior subtask history is included.

Overall, subtask annotation covers approximately 1.88 million episodes, 1.79 billion frames, and 20{,}101 hours. Automatic subtask annotation processes 78 million sampled frames (4.4\% of the total). From approximately 294{,}000 of the annotated episodes, we construct 1.1 million single-turn VL examples across 22 data families. Table~\ref{tab:subtask_sources} reports the source composition of the VL corpus.

\begin{table}[!htbp]
    \centering
    \reporttablestyle

    \caption{Subtask VL examples by source group. Native denotes source-provided subtask labels.}
    \label{tab:subtask_sources}
    \begin{tabular}{@{}llr@{\hspace{18pt}}llr@{}}
        \toprule
        Source Group & Labels & Examples & Source Group & Labels & Examples \\
        \midrule
        EgoDex        & VLM    & 125,000 & RoboCasa-pretrain & VLM    & 40,000 \\
        InternVLA-sim & VLM    & 103,415 & UniFoLM           & VLM    & 35,789 \\
        ABC-130K      & VLM    & 93,621  & Table30           & VLM    & 34,884 \\
        RoboCOIN      & Native & 91,642  & RW-RL             & VLM    & 26,397 \\
        OpenNeoData   & VLM    & 83,917  & Humanoid Everyday & VLM    & 24,866 \\
        GALAXEA       & Native & 75,022  & community-v3      & VLM    & 20,838 \\
        FastUMI       & VLM    & 70,597  & BC-Z              & VLM    & 20,000 \\
        RealOmin      & VLM    & 57,864  & RoboCasa-target   & Native & 18,000 \\
        RoboTwin2     & VLM    & 52,691  & E-Bench           & VLM    & 10,983 \\
        H-EmbodVis    & VLM    & 48,376  & community-v1      & VLM    & 8,675 \\
        Table30-v2    & VLM    & 43,862  & community-v2      & VLM    & 8,506 \\
        \midrule
        \multicolumn{5}{l}{Total} & 1,094,945 \\
        \bottomrule
    \end{tabular}
\end{table}

\subsubsection{ECoT Annotation}
Subtask labels identify the current intermediate goal but provide limited detail about the scene, task progress, and the motion that follows. We therefore augment the segmented trajectories in Section~\ref{sec:subtask-aug} with ECoT annotations~\citep{zawalski2024ecot,feng2026procvlm,sun2026ervla} that connect each subtask to its visual context and subsequent motion. Following Qwen-RobotManip~\citep{qwen2026robotmanip}, we annotate selected keyframes with scene descriptions, progress assessments, and next-motion descriptions derived from recorded poses.

\paragraph{Keyframe Selection.}
Keyframes are drawn only from frames marked valid during data processing (Section~\ref{sec:data_processing}). Within each subtask, we retain its first and last valid frame and select additional frames when their visual similarity to the previous keyframe falls below a threshold. We compute image hashes with the Python ImageHash library\footnote{\url{https://github.com/JohannesBuchner/imagehash}} to measure this similarity.

\paragraph{Scene and Progress Annotation.}
For each keyframe, Qwen3.8-27B describes task-relevant objects, spatial relations, and gripper state, and assesses whether the active subtask is complete. To account for earlier events, we uniformly sample frames between the episode's first frame and the current keyframe. Qwen3.8-27B uses these observations to generate a textual memory summary of gripper motion, object motion, and state changes. The model then uses the summary and the current frame to assess progress and explain its assessment. The summary is used during annotation and omitted from the training input.

\paragraph{Next-Motion Annotation.}
Where recorded EEF poses are available, we derive the next-motion description by comparing the EEF poses and gripper openings at the current keyframe and 0.5 seconds later. We ignore translation, rotation, and opening changes below their respective thresholds and map the remaining changes to directional phrases describing EEF motion and gripper opening or closing. These phrases are combined into a sentence such as ``\textit{The gripper moves forward, tilts down, and rotates clockwise.}''

We apply this pipeline to selected real-robot and UMI datasets, producing 2{,}564{,}851 annotated keyframes (Table~\ref{tab:pretrain_ecot}). The annotations yield both ECoT-annotated VLA trajectories and VL examples whose targets contain the scene description, progress assessment, and, where available, next-motion description. We additionally use the released annotations from ProcCorpus-60M~\citep{feng2026procvlm}.

\begin{table}[!htbp]
    \centering
    \reporttablestyle

    \caption{ECoT keyframes annotated by our pipeline.}
    \label{tab:pretrain_ecot}
    \begin{tabular}{@{}llr@{}}
        \toprule
        Dataset & Data Type & Annotated Keyframes \\
        \midrule
        AgiBot World~\citep{bu2025agibot} & Real & 530,301 \\
        ABC-130K~\citep{abc2026}          & Real & 530,289 \\
        RoboCOIN~\citep{wu2025robocoin}     & Real & 691,908 \\
        RoboMIND~\citep{robomind2025}     & Real & 153,508 \\
        HiFi-UMI-2K~\citep{simpleai2026hifiumi}      & UMI  & 575,197 \\
        FastUMI-100K~\citep{liu2025fastumi100k}       & UMI  & 83,648  \\
        \midrule
        Total                             & --   & 2,564,851 \\
        \bottomrule
    \end{tabular}
\end{table}

\FloatBarrier
\section{Model Architecture}

\begin{figure}[!htbp]
    \centering
    \includegraphics[width=\linewidth]{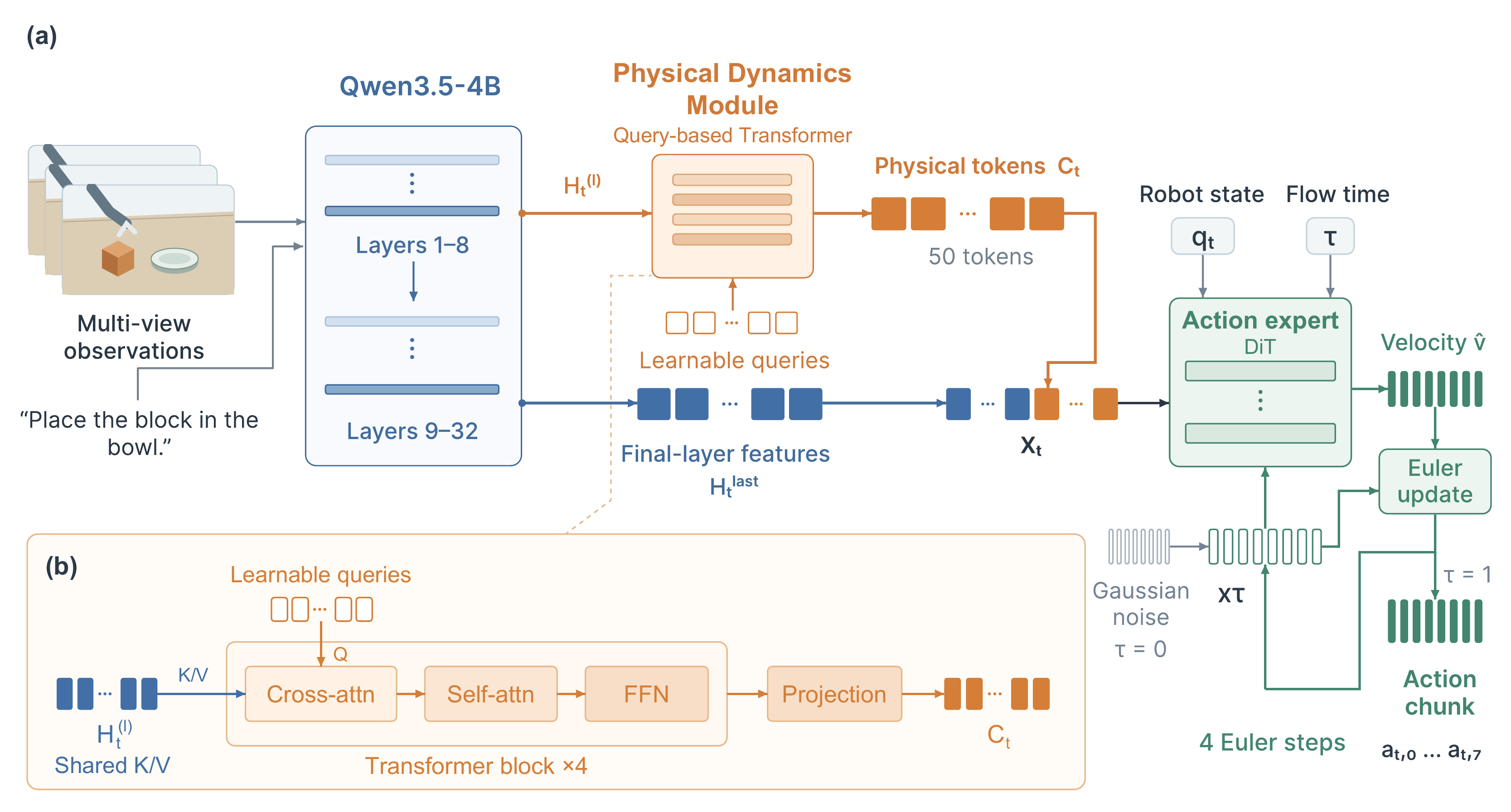}
    \caption{\textbf{\ourmodel{} architecture.} (a) Physical tokens complement final-layer VLM features as inputs to the action expert. Starting from Gaussian noise, Euler updates integrate predicted velocities into an action chunk. (b) Learnable queries attend to shared VLM features at every physical-module block before projection to physical tokens. Camera views and token sequences are schematic.}
    \label{fig:method}
\end{figure}

As shown in Figure~\ref{fig:method}, \ourmodel{} consists of a Qwen3.5-4B VLM backbone~\citep{qwen3.5}, the Physical Dynamics Module, and an action expert. Given visual observations and a task instruction, the backbone provides intermediate features to the physical module, which produces a fixed-length sequence of physical tokens. The action expert combines these tokens with the final-layer VLM features and the current proprioceptive state to generate a chunk of robot commands.

\FloatBarrier

\subsection{Input Construction}
\label{subsec:input_construction}

The VLM input is a single user message containing the current camera views followed by the task text, formatted with Qwen3.5's native chat template. Views follow the order specified in each dataset's camera configuration. We prefix the task instruction with the robot embodiment, arm and end-effector types, and control frequency when available, using the format \texttt{embodiment <robot>, arm <arm>, hand <hand>. control <rate> Hz. <instruction>}. For action generation with or without subtask conditioning, we use the VLM features directly without appending an assistant-generation prefix. The current proprioceptive state is supplied separately to the action expert.

Subtask descriptions can be appended to the task instruction as \texttt{Current step: <subtask>}. During training, this text comes from the trajectory annotations. At inference, \ourmodel{} can first generate a subtask description from the current observations and task instruction, then append it to the instruction for the action-conditioning pass. Without subtask conditioning, the policy generates actions directly from the current observations and original task instruction, together with the robot state.

For ECoT-conditioned action generation, the VLM first generates a response describing the scene, task progress, and next motion, which provides additional context to the action expert.

\subsection{Physical Dynamics Module}
\label{subsec:physical_module}

Later VLM layers may suppress visual information useful for predicting physical change. Layer-wise studies have found stronger perceptual representations at intermediate layers of large language models and a prominent role for these layers in accessing visual information~\citep{jain2025visper,kaduri2025image}. Motivated by these findings, at timestep $t$, we use the VLM's layer-$l$ features $H_t^{(l)}$ for the current multi-view observations and task instruction as input to the Physical Dynamics Module. In our implementation, $l=8$ for a backbone with 32 decoder blocks. The action expert also uses the final-layer features from the same backbone pass.

As shown in Figure~\ref{fig:method}(b), the module uses 50 learnable queries to extract a fixed set of physical tokens $C_t$ from $H_t^{(l)}$. Across 4 Transformer blocks, the queries retrieve information from the VLM features through cross-attention, exchange information through self-attention, and pass through a feed-forward layer. Each block has a hidden width of 1024, 4 attention heads, and a feed-forward expansion ratio of 1. A final projection maps the query outputs to the VLM hidden width, producing 50 physical tokens of width 2560. Future-prediction supervision trains the module to encode dynamics-related information in these tokens (Section~\ref{subsec:pretraining}), which are subsequently supplied to the action expert alongside the final-layer VLM features. The module adds approximately $57.3$\,M parameters ($1.4\%$ of the 4B backbone).

Appendix~\ref{app:physical_design} compares the VLM input layer, cross-attention placement, direct-supervision weighting, and number of learnable queries.

\subsection{Action Generation}
\label{subsec:action_policy}

At timestep $t$, the action policy receives the final-layer VLM features $H_t^{\text{last}}$, the physical tokens $C_t$, and a proprioceptive vector $q_t$ encoding the current robot state. For Cartesian control, the EEF pose in $q_t$ comprises three position coordinates and a continuous 6D orientation representation obtained by flattening the first two columns of its rotation matrix. The policy predicts a chunk $\mathbf{A}_t=[\mathbf{a}_{t,0},\ldots,\mathbf{a}_{t,K-1}]$ of $K=8$ robot commands, where $k$ in $\mathbf{a}_{t,k}$ indexes a step within the chunk predicted at $t$. The complete chunk is executed before the policy replans from a new observation.

\paragraph{Cross-Embodiment Action Representation.}
Training a single policy across heterogeneous robots requires a common representation that accommodates different controller interfaces. We use a fixed-dimensional action vector with separate left and right manipulator slots. Each slot reserves dimensions for four component types:
\begin{itemize}
    \item arm-joint positions: 7 dimensions;
    \item end-effector (EEF) pose: 9 reserved dimensions, of which 6 encode the action;
    \item dexterous-hand joint positions: 12 dimensions;
    \item gripper opening: 1 dimension.
\end{itemize}
The two slots together form a 58-dimensional vector. The EEF slot shares its reserved width with the 9D proprioceptive pose representation; its action occupies the first 6 dimensions, with the remaining 3 padded and masked. EEF actions are expressed relative to the pose at the beginning of the chunk, while arm-joint positions, dexterous-hand joint positions, and gripper openings are absolute targets. Dimensions unsupported by an embodiment's controller are padded and excluded from the training loss; execution uses only the supported components.

For Cartesian control, all EEF targets within a chunk are expressed relative to the EEF pose at the chunk's start. Let $\mathbf{p}_t\in\mathbb{R}^3$ and $R_t\in SO(3)$ denote this starting position and orientation, and let $\mathbf{p}^*_{t+k}$ and $R^*_{t+k}$ denote the target position and orientation for step $k=0,\ldots,K-1$. The superscript $*$ denotes a target. Omitting the left/right manipulator index, each 6D EEF action comprises a translation delta and a rotation delta:
\begin{equation}
    \mathbf{a}^{\mathrm{eef}}_{t,k}
    = \left[\mathbf{p}^*_{t+k}-\mathbf{p}_t,\;
    \operatorname{Log}\!\left(R^*_{t+k}R_t^{\mathsf T}\right)\right]
    \in \mathbb{R}^{6},
    \label{eq:anchor_delta}
\end{equation}
where $\operatorname{Log}:SO(3)\rightarrow\mathbb{R}^3$ maps the relative rotation to an axis--angle vector. Target poses are obtained as described in Section~\ref{sec:data_processing}.

\paragraph{Action Expert.}
We train a Diffusion Transformer (DiT)-based action expert with conditional flow matching. It receives a noisy action chunk, whose action vectors are mapped to action tokens by an input projection shared across embodiments. We concatenate the final-layer VLM features and physical tokens along the token dimension to form the conditioning sequence:
\begin{equation}
    X_t = \left[ H_t^{\text{last}}; \, C_t \right],
    \label{eq:dual_path_fusion}
\end{equation}

The expert alternates self-attention over its input tokens with cross-attention to $X_t$. The current proprioceptive state $q_t$ is encoded as a prefix token, while flow time conditions the action-token encoding and the Transformer's adaptive layer normalization. A shared output projection maps the resulting action-token features to a velocity field over the chunk. The action expert contains approximately $150$\,M parameters. At inference, we initialize the chunk with Gaussian noise and integrate the predicted velocity field from flow time 0 to 1 using four Euler steps. Section~\ref{sec:training} defines the interpolation path and masked velocity-matching loss.

\section{Training}
\label{sec:training}

\begin{figure}[!htbp]
    \centering
    \includegraphics[width=\linewidth]{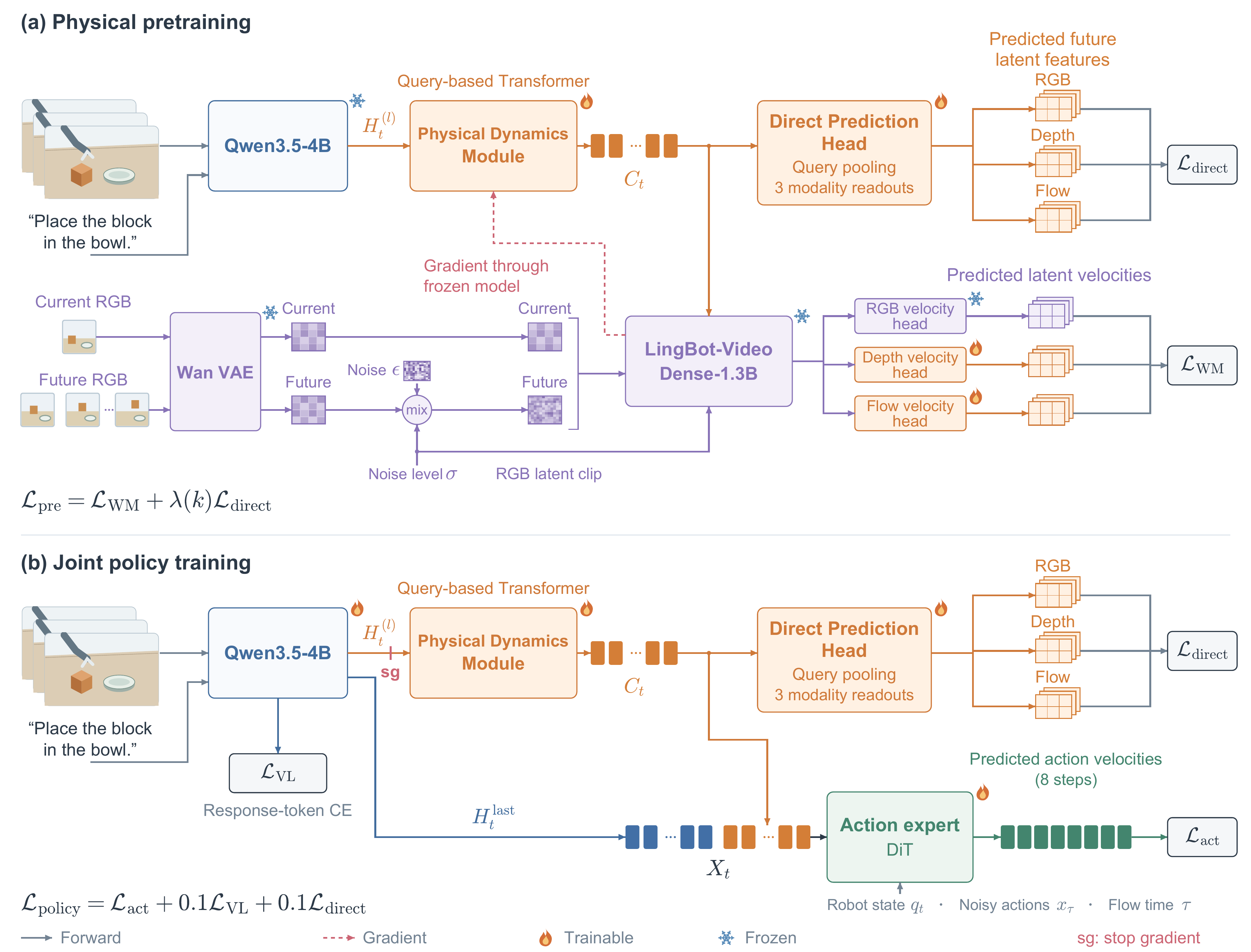}
    \caption{\textbf{Training stages.} (a) Physical pretraining combines direct future-feature prediction with supervision through a frozen video model. (b) Joint policy training removes the video model, retains direct prediction, and adds action and VL objectives. Grid stacks represent feature tensors; \emph{sg} marks the detached intermediate-feature path. Camera views and video frames are schematic.}
    \label{fig:training_framework}
\end{figure}

Figure~\ref{fig:training_framework} summarizes physical-module pretraining and joint policy training. The resulting policy is then adapted to each benchmark. Physical pretraining combines direct prediction of future RGB, depth, and optical-flow latent features with auxiliary supervision through a frozen pretrained video model. We then jointly train the VLM, Physical Dynamics Module, and action expert on VLA and VL data, retaining the three direct prediction losses alongside action and VL supervision. The video world model and its output heads are used only during physical pretraining. Additional architecture settings, optimization parameters, and sampling details are provided in Appendix~\ref{app:training_details}.

\subsection{Physical Dynamics Pretraining}
\label{subsec:pretraining}

We pretrain the Physical Dynamics Module (Section~\ref{subsec:physical_module}) to encode future appearance, geometry, and motion in its physical tokens. A direct prediction objective supervises these tokens using future RGB, depth, and optical-flow latent features. Because the module and its prediction head are trained from scratch, we also use a frozen pretrained video model to provide auxiliary supervision for representation learning.

\paragraph{Future Prediction Targets.}
For physical pretraining, each example pairs the current observations and task instruction with a five-frame video sampled from a recorded clip spanning the current timestep and the following 30 action steps. The RGB frames provide appearance targets, while DA3~\citep{lin2025depth} and DPFlow~\citep{morimitsu2025dpflow} generate depth and optical-flow pseudo-labels for geometry and motion supervision. Following RynnWorld-4D~\citep{zhao2026rynnworld4d}, depth values are clipped to a fixed range and mapped to grayscale intensities, while optical-flow fields are color-encoded to form the corresponding depth and flow videos.
We encode the three video targets into compact latent features rather than predicting them in pixel space. LingBot-Video's frozen Wan video variational autoencoder (VAE)~\citep{wan2025} provides these representations, and the same encoded targets are reused for direct prediction.

\paragraph{Direct Future Prediction.}
The direct prediction head applies layer normalization and a linear projection to the physical tokens $C_t$, then uses learned spatial queries for scaled dot-product attention pooling. Three modality-specific linear readouts predict future RGB, depth, and optical-flow features. Let $z_{\mathrm{future}}^m$ denote the future portion of the VAE-encoded video for modality $m\in\{r,d,f\}$, corresponding to RGB, depth, and optical flow. We apply adaptive average pooling to obtain $y^m=\operatorname{Pool}(z_{\mathrm{future}}^m)$. This reduces the spatial resolution of the prediction target while retaining a coarse spatial layout, so the lightweight head need not reconstruct the full-resolution latent features.

To reduce differences in target scale across modalities and feature dimensions, we standardize the pooled targets as
\begin{equation}
    \widetilde y^m=\frac{y^m-\mu^m}{s^m},
    \label{eq:direct_target_standardization}
\end{equation}
where $\mu^m$ and $s^m$ are estimated from training samples separately for each modality, channel, and spatial position. A lower bound on the standard deviation prevents excessive amplification of low-variance features. Denoting the head's prediction for modality $m$ by $g_\eta^m(C_t)$, the direct prediction loss is
\begin{equation}
    \mathcal{L}_{\mathrm{direct}} = \frac{1}{3} \sum_{m \in \{r, d, f\}} \operatorname{MSE}\left(g_\eta^m(C_t), \, \widetilde y^m\right).
    \label{eq:direct_loss}
\end{equation}

\paragraph{World-Model-Assisted Prediction.}
Beyond direct feature regression, we use a pretrained video model to guide physical representation learning, following related video-conditioning approaches~\citep{cai2026internvlaa15,tang2026worldtokens}. The frozen LingBot-Video-Dense-1.3B model~\citep{ma2026lingbotvideo} receives an RGB latent clip, the physical tokens $C_t$ as conditioning inputs, and a noise timestep. Future-prediction losses train the Physical Dynamics Module to provide context for this video predictor. In addition to its frozen RGB output head, two trainable heads read LingBot's final hidden features to predict depth and optical-flow velocities, extending the auxiliary supervision to future geometry and motion.

Within the RGB latent clip, the current features remain clean and the future features are mixed with Gaussian noise. Let $z_{\mathrm{current}}^r$ and $z_{\mathrm{future}}^r$ denote these two parts of the encoded clip. We sample a noise level $\sigma\sim\mathcal{U}(0,1)$ and Gaussian noise $\epsilon\sim\mathcal{N}(0,I)$ with the same shape as the future latent. The RGB input is constructed as
\begin{equation}
    z^r_{\sigma,\mathrm{current}}=z_{\mathrm{current}}^r,
    \qquad
    z^r_{\sigma,\mathrm{future}}
    =(1-\sigma)z_{\mathrm{future}}^r+\sigma\epsilon.
    \label{eq:fm_latent}
\end{equation}

The three output heads predict future latent velocities $\widehat v_{\mathrm{future}}^m$ for $m\in\{r,d,f\}$. For the linear data-to-noise path in Eq.~\ref{eq:fm_latent}, the target velocity is the difference between the noise and the clean latent. We use the same noise tensor $\epsilon$ for all three modalities, yielding targets $\epsilon-z_{\mathrm{future}}^m$. The auxiliary loss sums the three equally weighted velocity errors:
\begin{equation}
    \mathcal{L}_{\mathrm{WM}} = \sum_{m \in \{r,d,f\}}
    \operatorname{MSE}\!\left(\widehat v_{\mathrm{future}}^m,
    \epsilon-z_{\mathrm{future}}^m\right).
    \label{eq:wm_loss}
\end{equation}
Gradients from $\mathcal{L}_{\mathrm{WM}}$ train the depth and optical-flow heads and pass through the frozen LingBot backbone to train the Physical Dynamics Module. The RGB head and LingBot backbone are not updated.

\paragraph{Pretraining Objective.}
The complete physical-pretraining objective combines direct future prediction with the auxiliary flow-matching loss:
\begin{equation}
    \mathcal{L}_{\mathrm{pre}} = \mathcal{L}_{\mathrm{WM}} + \lambda(k) \mathcal{L}_{\mathrm{direct}},
    \label{eq:total_pretrain_loss}
\end{equation}
where $k$ denotes the optimization step and $\lambda(k)=0.5\min(k/2000,1)$ increases the direct prediction weight linearly from 0 to 0.5 over the first 2000 steps, after which it remains constant.

Physical pretraining updates the Physical Dynamics Module, the depth and optical-flow output heads, and the direct prediction head. The VLM and the pretrained LingBot-Video components remain frozen.

\FloatBarrier
\subsection{Action Policy Training}
\label{subsec:policy_adaptation}

After physical pretraining, we attach the action expert and jointly train the complete policy on VLA examples and the general, spatial, embodied, ECoT, and subtask VL data. The direct prediction head is retained to supervise future RGB, depth, and optical-flow features during action learning. The video world model and its RGB, depth, and optical-flow output heads are no longer used.

\paragraph{Joint Policy Training.}
The VLM, pretrained Physical Dynamics Module, action expert, and direct prediction head are all trainable. Physical tokens are concatenated with the final-layer VLM features to form $X_t=[H_t^{\text{last}};C_t]$ for the action expert (Section~\ref{subsec:action_policy}). The intermediate VLM features entering the physical module are detached: action and direct prediction losses update the physical module but do not propagate into the VLM through this input. The action loss updates the VLM through the final-layer conditioning path, while the direct losses also update their prediction head. VL supervision updates only the VLM.

\paragraph{Training Objectives.}
Conditional flow matching trains the action expert to transform Gaussian noise into an action chunk. Given a target chunk $\mathbf{A}_t$ containing $K=8$ actions, we sample noise $\xi\sim\mathcal{N}(0,I)$ of the same shape and a flow time $\tau\in[0,1]$. The interpolated chunk and target velocity are
\begin{equation}
    x_\tau = (1-\tau)\xi + \tau \mathbf{A}_t,
    \qquad v^* = \mathbf{A}_t - \xi.
    \label{eq:action_flow_path}
\end{equation}
The action expert $v_\psi(x_\tau,\tau,X_t,q_t)$ predicts the velocity from the interpolated chunk, flow time, conditioning features $X_t=[H_t^{\text{last}};C_t]$, and current proprioceptive state.

Frames marked invalid during data processing (Section~\ref{sec:data_processing}) are excluded as starting observations for training samples. Within a retained chunk, action steps whose target frames are invalid are excluded from the loss, as are padded steps beyond the episode boundary. For each sample $i$, a two-dimensional binary supervision mask $M_i$ has rows corresponding to chunk steps and columns to action-vector dimensions. An entry is 1 only if the step is valid and the dimension is supported by the robot's controller; otherwise, it is 0. The masked velocity-matching loss over a minibatch $\mathcal{B}$ is
\begin{equation}
    \mathcal{L}_{\mathrm{act}} = \frac{\sum_{i\in\mathcal{B}}\| M_i \odot (\widehat{v}_i-v_i^*) \|_F^2}{\sum_{i\in\mathcal{B}}\| M_i \|_1 + \delta},
    \label{eq:masked_action_loss}
\end{equation}
where $\widehat{v}_i$ and $v_i^*$ are the action expert's predicted velocity and the target velocity for sample $i$, respectively, and $\odot$ denotes element-wise multiplication. The loss is averaged over all supervised action elements in the minibatch, with $\delta=10^{-6}$ preventing division by zero.

Alongside action prediction, VL examples supervise the VLM through next-token prediction of target responses. The VL loss $\mathcal{L}_{\mathrm{VL}}$ is the cross-entropy on response tokens only, averaged first within each example and then across examples. It updates only the VLM, without passing through the action expert or Physical Dynamics Module.

For VLA examples with future targets, $\mathcal{L}_{\mathrm{direct}}$ averages the retained direct prediction head's RGB, depth, and optical-flow prediction errors, as defined in Eq.~\ref{eq:direct_loss}. Target construction follows Section~\ref{subsec:pretraining}, including pooling and the standardization in Eq.~\ref{eq:direct_target_standardization}. Each optimization step combines VLA and VL supervision with these direct prediction losses:
\begin{equation}
    \mathcal{L}_{\mathrm{policy}}=\mathcal{L}_{\mathrm{act}}+0.1\mathcal{L}_{\mathrm{VL}}+0.1\mathcal{L}_{\mathrm{direct}}.
    \label{eq:policy_joint_loss}
\end{equation}

\paragraph{VLA Frame Sampling.}
To reduce closely spaced training examples while retaining changes in end-effector control, we select chunk starting frames from a regular temporal grid at approximately 8~Hz and from frames with changes in gripper or dexterous-hand target commands. Selection is performed when building the training index. Each selected frame anchors an eight-step action chunk read from the original, temporally aligned stream without further subsampling.

\section{Experiments}
\label{sec:experiments}
We evaluate \ourmodel{} on four simulation benchmarks covering single-arm and bimanual manipulation, robustness to observation and instruction perturbations, and WidowX manipulation under visual-matching settings. Performance is reported as task success rates and compared with published VLA baselines. Section~\ref{sec:ablation_analysis} examines the effect of world-model-assisted pretraining on future prediction. Additional studies examine subtask conditioning and its update frequency, auxiliary future object-pose supervision, and ECoT supervision during training versus generation at inference.

\subsection{Experimental Setup}

LIBERO~\cite{liu2023libero} evaluates single-arm manipulation across four suites: Spatial, Object, Goal, and Long, each containing 10 tasks. We fine-tune the pretrained policy jointly on LIBERO demonstrations from all four suites for 40{,}000 updates with a global batch size of 128. Adaptation uses dense valid starting frames rather than the frame-selection procedure used during policy pretraining. We evaluate 500 episodes per suite (2{,}000 in total) and report success rates for each suite and their average.

LIBERO-Plus~\cite{fei2025liberoplus} evaluates the LIBERO tasks under camera, robot, language, lighting, background, noise, and layout perturbations. We evaluate the LIBERO checkpoint and report success rates for each perturbation category and across all 10{,}030 evaluation cases.

RoboTwin 2.0~\cite{chen2025robotwin} evaluates bimanual manipulation on 50 Aloha-AgileX tasks. We fine-tune the pretrained policy for 150{,}000 updates with a global batch size of 1024, using the RoboTwin 2.0 Aloha-AgileX \texttt{clean50} and \texttt{randomized500} demonstration sets. Valid starting frames are sampled densely. We evaluate the resulting checkpoint for 100 episodes per task in each setting and report clean and randomized success rates together with their average.

SimplerEnv~\cite{li2025simpler} evaluates WidowX manipulation on four Bridge visual-matching tasks. We fine-tune the pretrained policy on the full Bridge training set for 50{,}000 updates with a global batch size of 128, using dense valid starting frames. We report per-task success rates and their average.

\FloatBarrier
\subsection{Main Results}

Across the four simulation benchmarks, \ourmodel{} achieves the highest aggregate success rates among the compared methods, demonstrating strong manipulation performance and robustness under benchmark perturbations. Baseline results are drawn from the cited papers and official repositories under their respective training and evaluation protocols, including comparisons reported by InternVLA-A1.5~\citep{cai2026internvlaa15} and Qwen-RobotManip~\citep{qwen2026robotmanip}. Table~\ref{tab:sim_overview} summarizes the aggregate results, followed by detailed comparisons for each benchmark.

\begin{table}[!htbp]
    \centering
    \reporttablestyle

    \caption{Overview of simulation results (success rate, \%). \NA{} denotes unavailable results.}
    \label{tab:sim_overview}
    \begin{tabular}{@{}lrrrr@{}}
    \toprule
    Method & LIBERO & LIBERO-Plus & RoboTwin 2.0 & SimplerEnv \\
    \midrule
    $\pi_0$~\cite{black2024pi0} & 94.4 & 53.6 & 62.2 & 27.1 \\
    $\pi_{0.5}$~\cite{intelligence2025pi05} & 96.9 & 84.4 & 79.8 & 57.1 \\
    OpenVLA-OFT~\cite{kim2025openvlaoft} & 97.1 & 69.6 & \NA & \NA \\
    GR00T-N1.7~\cite{bjorck2025gr00t,nvidia2026gr00tsimpler} & 97.0 & \NA & \NA & 59.3 \\
    Cosmos-Policy~\cite{kim2026cosmospolicy} & 98.5 & 82.2 & \NA & \NA \\
    LingBot-VLA 1.0~\cite{lingbotvla2026} & \NA & \NA & 87.6 & \NA \\
    LingBot-VLA 2.0~\cite{lingbotvlav22026} & \NA & \NA & 93.2 & \NA \\
    LingBot-VA~\cite{lingbotva2026} & 98.5 & \NA & 92.3 & \NA \\
    Xiaomi-Robotics-0~\cite{cai2026xiaomi} & 98.7 & \NA & \NA & 79.2 \\
    Qwen-VLA-Instruct~\cite{qwen2026vla} & 97.9 & \NA & 86.7 & 73.7 \\
    Qwen-RobotManip~\cite{qwen2026robotmanip} & 99.1 & 89.0 & 93.0 & \NA \\
    InternVLA-A1.5~\cite{cai2026internvlaa15} & 98.9 & 84.8 & 93.2 & 80.8 \\
    \rowcolor{ourrow}
    \ourmodel{} & \textbf{99.3} & \textbf{89.9} & \textbf{93.6} & \textbf{81.5} \\
    \bottomrule
    \end{tabular}

\end{table}
\FloatBarrier

\paragraph{LIBERO.}
On LIBERO, \ourmodel{} achieves the highest average success rate among the compared methods at 99.3\%, compared with 99.1\% for Qwen-RobotManip and 98.9\% for InternVLA-A1.5 (Table~\ref{tab:libero}). It achieves 100.0\% success on Spatial and 99.8\% on both Object and Goal. On Long, its success rate is 97.6\%, below LingBot-VA's 98.5\%.

\begin{table}[!htbp]
    \centering
    \reporttablestyle

    \caption{LIBERO success rates (\%).}
    \label{tab:libero}
    \begin{tabular}{@{}lrrrrr@{}}
    \toprule
    Method & Spatial & Object & Goal & Long & Avg. \\
    \midrule
    $\pi_0$~\cite{black2024pi0} & 98.0 & 96.8 & 94.4 & 88.4 & 94.4 \\
    $\pi_{0.5}$~\cite{intelligence2025pi05} & 98.8 & 98.2 & 98.0 & 92.4 & 96.9 \\
    OpenVLA-OFT~\cite{kim2025openvlaoft} & 97.6 & 98.4 & 97.9 & 94.5 & 97.1 \\
    GR00T-N1.7~\cite{bjorck2025gr00t} & 97.7 & 98.5 & 97.5 & 94.4 & 97.0 \\
    Cosmos-Policy~\cite{kim2026cosmospolicy} & 98.1 & \textbf{100.0} & 98.2 & 97.6 & 98.5 \\
    LingBot-VA~\cite{lingbotva2026} & 98.5 & 99.6 & 97.2 & \textbf{98.5} & 98.5 \\
    Xiaomi-Robotics-0~\cite{cai2026xiaomi} & 98.8 & \textbf{100.0} & 98.8 & 97.2 & 98.7 \\
    Qwen-VLA-Instruct~\cite{qwen2026vla} & \NA & \NA & \NA & \NA & 97.9 \\
    Qwen-RobotManip~\cite{qwen2026robotmanip} & \NA & \NA & \NA & \NA & 99.1 \\
    InternVLA-A1.5~\cite{cai2026internvlaa15} & 98.6 & 99.8 & 98.6 & 98.4 & 98.9 \\
    \rowcolor{ourrow}
    \ourmodel{} & \textbf{100.0} & 99.8 & \textbf{99.8} & 97.6 & \textbf{99.3} \\
    \bottomrule
    \end{tabular}

\end{table}
\FloatBarrier

\paragraph{LIBERO-Plus.}
On LIBERO-Plus, \ourmodel{} achieves the highest aggregate success rate among the compared methods at 89.9\%, compared with 89.0\% for Qwen-RobotManip and 84.8\% for InternVLA-A1.5 (Table~\ref{tab:libero-plus}). It leads in four of the seven perturbation categories, while Qwen-RobotManip performs better under robot-initial-state, sensor-noise, and object-layout changes. Robot-initial-state perturbations are the most challenging category for \ourmodel{}, with a success rate of 75.2\%.

\begin{table}[!htbp]
    \centering
    \reporttablestyle

    \caption{LIBERO-Plus success rates by perturbation category (\%).}
    \label{tab:libero-plus}

    \setlength{\tabcolsep}{2.5pt}
    \begin{tabular}{@{}lrrrrrrrr@{}}
    \toprule
    Method & \makecell{Sensor\\noise} & \makecell{Camera\\viewpoints} & \makecell{Robot\\initial states} & \makecell{Light\\conditions} & \makecell{Language\\instructions} & \makecell{Background\\textures} & \makecell{Object\\layout} & Overall \\
    \midrule
    $\pi_0$~\cite{black2024pi0} & 79.0 & 13.8 & 6.0 & 85.0 & 58.8 & 81.4 & 68.9 & 53.6 \\
    $\pi_{0.5}$~\cite{intelligence2025pi05} & 89.0 & 78.4 & 73.6 & 96.2 & 80.8 & 94.1 & 84.5 & 84.4 \\
    OpenVLA-OFT~\cite{kim2025openvlaoft} & 75.8 & 56.4 & 31.9 & 88.7 & 79.5 & 93.3 & 74.2 & 69.6 \\
    Cosmos-Policy~\cite{kim2026cosmospolicy} & 92.7 & 75.8 & 63.3 & 96.5 & 81.7 & 88.9 & 82.2 & 82.2 \\
    Qwen-RobotManip~\cite{qwen2026robotmanip} & \textbf{97.7} & 87.2 & \textbf{75.5} & 96.6 & 85.6 & 97.7 & \textbf{87.3} & 89.0 \\
    InternVLA-A1.5~\cite{cai2026internvlaa15} & 95.6 & 83.1 & 55.1 & 96.4 & 86.9 & 98.2 & 85.2 & 84.8 \\
    \rowcolor{ourrow}
    \ourmodel{} & 95.6 & \textbf{88.2} & 75.2 & \textbf{99.1} & \textbf{90.7} & \textbf{99.4} & 86.5 & \textbf{89.9} \\
    \bottomrule
    \end{tabular}

\end{table}
\FloatBarrier

\paragraph{RoboTwin 2.0.}
On RoboTwin 2.0, \ourmodel{} achieves the highest average success rate among the compared methods at 93.6\%, compared with 93.2\% for both LingBot-VLA 2.0 and InternVLA-A1.5 and 93.0\% for Qwen-RobotManip (Table~\ref{tab:robotwin}). It achieves 93.9\% success in the clean setting and 93.3\% in the randomized setting, leading the compared methods in both settings.

\begin{table}[!htbp]
    \centering
    \reporttablestyle

    \caption{RoboTwin 2.0 success rates (\%) in clean and randomized settings and their average.}
    \label{tab:robotwin}
    \begin{tabular}{@{}lrrr@{}}
    \toprule
    Method & Clean & Rand. & Avg. \\
    \midrule
    $\pi_0$~\cite{black2024pi0} & 65.9 & 58.4 & 62.2 \\
    $\pi_{0.5}$~\cite{intelligence2025pi05} & 82.7 & 76.8 & 79.8 \\
    LingBot-VLA 1.0~\cite{lingbotvla2026} & 88.6 & 86.7 & 87.6 \\
    LingBot-VLA 2.0~\cite{lingbotvlav22026} & 93.5 & 92.8 & 93.2 \\
    LingBot-VA~\cite{lingbotva2026} & 92.9 & 91.6 & 92.3 \\
    Qwen-VLA-Instruct~\cite{qwen2026vla} & 86.1 & 87.2 & 86.7 \\
    Qwen-RobotManip~\cite{qwen2026robotmanip} & 93.4 & 92.5 & 93.0 \\
    InternVLA-A1.5~\cite{cai2026internvlaa15} & 93.3 & 93.0 & 93.2 \\
    \rowcolor{ourrow}
    \ourmodel{} & \textbf{93.9} & \textbf{93.3} & \textbf{93.6} \\
    \bottomrule
    \end{tabular}

\end{table}
\FloatBarrier

\paragraph{SimplerEnv.}
On SimplerEnv, \ourmodel{} achieves the highest average success rate among the compared methods at 81.5\%, compared with 80.8\% for InternVLA-A1.5 and 79.2\% for Xiaomi-Robotics-0 (Table~\ref{tab:simpler}). Its strongest result is on Eggplant$\to$Basket, where it achieves 97.6\% success, compared with 83.3\% for Xiaomi-Robotics-0 and 75.7\% for InternVLA-A1.5. On Carrot$\to$Plate, its success rate is 64.6\%, below InternVLA-A1.5's 85.4\%.

\begin{table}[!htbp]
    \centering
    \reporttablestyle

    \caption{SimplerEnv (WidowX / Bridge, visual matching) success rates (\%).}
    \label{tab:simpler}

    \begin{tabular}{@{}lrrrrr@{}}
    \toprule
    Method & \makecell{Spoon\\$\to$Towel} & \makecell{Carrot\\$\to$Plate} & \makecell{Stack\\Blocks} & \makecell{Eggplant\\$\to$Basket} & Avg. \\
    \midrule
    $\pi_0$~\cite{black2024pi0} & 29.1 & 0.0 & 16.6 & 62.5 & 27.1 \\
    $\pi_{0.5}$~\cite{intelligence2025pi05} & 49.3 & 64.7 & 44.7 & 69.7 & 57.1 \\
    GR00T-N1.7~\cite{nvidia2026gr00tsimpler} & 78.0 & 58.0 & 48.0 & 53.0 & 59.3 \\
    Xiaomi-Robotics-0~\cite{cai2026xiaomi} & \textbf{95.8} & 62.5 & \textbf{75.0} & 83.3 & 79.2 \\
    Qwen-VLA-Instruct~\cite{qwen2026vla} & \NA & \NA & \NA & \NA & 73.7 \\
    InternVLA-A1.5~\cite{cai2026internvlaa15} & 92.4 & \textbf{85.4} & 69.5 & 75.7 & 80.8 \\
    \rowcolor{ourrow}
    \ourmodel{} & 92.4 & 64.6 & 71.2 & \textbf{97.6} & \textbf{81.5} \\
    \bottomrule
    \end{tabular}

\end{table}
\FloatBarrier

\subsection{Component Analysis}
\label{sec:ablation_analysis}

We first examine whether world-model supervision improves future-feature prediction, then study the use of subtask and ECoT annotations in action policies. An additional exploratory experiment tests object-pose prediction as auxiliary supervision. The policy studies use independent training setups and baselines.

\subsubsection{Physical Representation Learning}
\begin{figure}[!htbp]
    \centering
    \includegraphics[width=\linewidth]{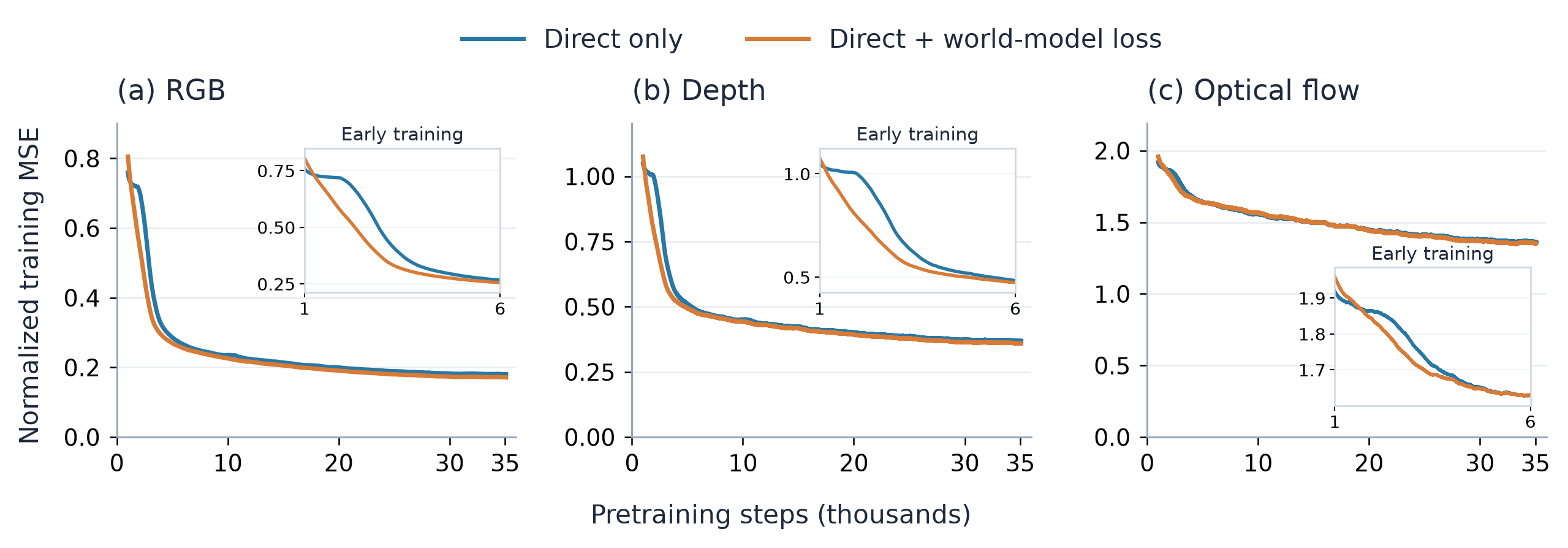}
    \caption{Normalized training MSE with direct prediction alone or additional world-model supervision. Each curve shows sample-weighted means over the preceding 1{,}000 updates; insets show updates 1{,}000--6{,}000 in greater detail.}
    \label{fig:physical_wm_ablation}
\end{figure}

We examine whether supervision from a video world model improves the physical module's ability to predict future features beyond direct supervision alone. We compare direct future-feature prediction with the same objective augmented by world-model-assisted flow matching. Both runs use the original Qwen3.5-4B backbone, equal training-sample budgets, and identical learning-rate and direct-loss-weight schedules. We compare their normalized RGB, depth, and optical-flow prediction errors before loss weighting, excluding the auxiliary world-model loss. Figure~\ref{fig:physical_wm_ablation} shows that RGB and depth training errors decrease more rapidly early in training with world-model assistance. The two runs show a smaller gap in optical-flow error, and the gaps narrow later in training.

To evaluate prediction on unseen data, we assess both runs on the same 828 RoboCOIN clips from 100 episodes excluded from training, using their original trained direct heads without refitting. Table~\ref{tab:physical_wm_endpoint} reports the final-checkpoint errors. The mean gives equal weight to the three modalities. Mean error decreases from 0.6439 to 0.6354, a relative reduction of 1.32\%, with the largest gains in RGB (4.06\%) and depth (2.12\%). The flow reduction is smaller (0.62\%), and its 95\% paired episode-bootstrap interval includes zero. Together, the training curves and held-out results show improved future-feature prediction, with the clearest benefits in RGB and depth.

\begin{table}[!htbp]
    \centering
    \reporttablestyle

    \caption{Final normalized prediction MSE on held-out RoboCOIN clips. Mean denotes the equally weighted average across the three modalities. Relative reductions are measured against Direct only; intervals are 95\% paired episode-bootstrap intervals.}
    \label{tab:physical_wm_endpoint}
    \begin{tabular}{@{}lrrrr@{}}
    \toprule
    Target & Direct only & Direct + WM & Reduction (\%) & 95\% interval (\%) \\
    \midrule
    RGB   & 0.2558 & \textbf{0.2454} & 4.06 & [3.04, 4.99] \\
    Depth & 0.3154 & \textbf{0.3087} & 2.12 & [0.93, 3.31] \\
    Flow  & 1.3606 & \textbf{1.3522} & 0.62 & [$-0.07$, 1.28] \\
    Mean  & 0.6439 & \textbf{0.6354} & 1.32 & [0.71, 1.89] \\
    \bottomrule
    \end{tabular}
\end{table}

\FloatBarrier
\subsubsection{Auxiliary Supervision for Action Policies}

\paragraph{Subtask Conditioning.}
We first test whether self-generated subtask descriptions improve action success on LIBERO-Plus~\citep{fei2025liberoplus}. At inference, the policy's own VLM predicts the current subtask from the observations and task instruction, then appends the description to the instruction for action generation. Table~\ref{tab:subtask_ablation} compares this setting with the same SFT checkpoint evaluated without preliminary subtask prediction.

\begin{table}[!htbp]
    \centering
    \reporttablestyle

    \caption{Self-generated subtask conditioning on LIBERO-Plus by perturbation category (success rate, \%).}
    \label{tab:subtask_ablation}
    \setlength{\tabcolsep}{2.5pt}
    \begin{tabular}{@{}lrrrrrrrr@{}}
    \toprule
    Variant & \makecell{Sensor\\noise} & \makecell{Camera\\viewpoints} & \makecell{Robot\\initial states} & \makecell{Light\\conditions} & \makecell{Language\\instructions} & \makecell{Background\\textures} & \makecell{Object\\layout} & Overall \\
    \midrule
    SFT baseline & 95.57 & 88.18 & 75.16 & \textbf{99.12} & \textbf{90.70} & \textbf{99.44} & 86.49 & 89.93 \\
    \makecell[l]{+ Subtask\\conditioning} & \textbf{96.50} & \textbf{89.18} & \textbf{76.32} & 98.51 & 89.92 & 99.07 & \textbf{86.75} & \textbf{90.23} \\
    \bottomrule
    \end{tabular}
\end{table}

Subtask conditioning raises overall success from 89.93\% to 90.23\%. Its largest gains occur under robot-initial-state and camera-viewpoint perturbations ($+1.16$ and $+1.00$ percentage points), with additional improvements under sensor noise and object-layout changes. Success decreases under lighting, language, and background-texture perturbations.

\paragraph{Subtask Update Frequency.}
We examine how often self-generated subtask descriptions need to be refreshed during execution on eight RoboCasa365~\citep{robocasa365} tasks, with 100 evaluation episodes per task under a fixed-seed setting. The policy's VLM predicts the current subtask from the observations and task instruction. We compare refresh intervals of 1, 8, and 32 action chunks using the same subtask-conditioned checkpoint. Between refreshes, the description is reused while actions are generated from updated observations. A separately trained policy without subtask conditioning serves as the baseline (Table~\ref{tab:subtask_conditioning}).

\begin{table}[!htbp]
    \centering
    \reporttablestyle

    \caption{RoboCasa365 success rates (\%) with different subtask refresh intervals.}
    \label{tab:subtask_conditioning}
    \begin{tabular}{@{}lrrrr@{}}
    \toprule
    & & \multicolumn{3}{c}{Subtask refresh interval (chunks)} \\
    \cmidrule(lr){3-5}
    Task & Baseline & 1 & 8 & 32 \\
    \midrule
    CuttingToolSelection & 46 & \textbf{60} & 50 & 33 \\
    RinseSinkBasin       & 39 & \textbf{49} & 44 & 37 \\
    PrepareCoffee        & 7  & \textbf{13} & 7  & 7  \\
    PanTransfer          & 22 & \textbf{25} & 21 & 15 \\
    LoadDishwasher       & 48 & 55 & \textbf{58} & 33 \\
    ArrangeTea           & 7  & 15 & \textbf{22} & 16 \\
    PreSoakPan           & 31 & 29 & \textbf{38} & 21 \\
    WashFruitColander    & 10 & 12 & \textbf{17} & 3  \\
    \midrule
    Average              & 26.25 & \textbf{32.25} & 32.13 & 20.63 \\
    \bottomrule
    \end{tabular}
\end{table}

Using refresh intervals of 1 or 8 chunks yields average success rates of 32.25\% and 32.13\%, respectively, compared with 26.25\% for the baseline. These similar averages mask task-level differences: changing the interval from 1 to 8 reduces CuttingToolSelection success from 60\% to 50\%, but increases ArrangeTea success from 15\% to 22\%. Extending the interval to 32 reduces average success to 20.63\%, below the baseline. An eight-chunk interval therefore gives similar aggregate success to per-chunk refreshes in this evaluation.

To characterize these updates, we measure the subtask change rate, defined as the percentage of successive predictions that differ. The rates are 11.6\%, 26.5\%, and 56.1\% for refresh intervals of 1, 8, and 32 chunks, respectively. Most per-chunk predictions therefore repeat the preceding subtask.

\paragraph{ECoT Training and Generation.}
We examine whether ECoT supervision improves action success and whether its benefits require ECoT generation at inference. We compare a baseline trained without ECoT with a policy whose VLM is trained to generate ECoT responses that provide context to the action expert. The ECoT-trained checkpoint is evaluated both with and without generation at inference (Table~\ref{tab:ecot_analysis}).

\begingroup
\setlength{\intextsep}{6pt}
\begin{table}[H]
    \centering
    \reporttablestyle
    \setlength{\belowcaptionskip}{5pt}
    \caption{Effects of ECoT training and inference-time generation on success rates (\%).}
    \label{tab:ecot_analysis}
    \begin{tabular}{@{}lrrrr@{}}
    \toprule
    & \multicolumn{2}{c}{LIBERO} & \multicolumn{2}{c}{LIBERO-Plus} \\
    \cmidrule(lr){2-3}\cmidrule(lr){4-5}
    Setting & Long & Average & Long & Overall \\
    \midrule
    No ECoT training & 92.8 & 97.1 & 65.18 & \textbf{79.59} \\
    ECoT-trained, no generation & \textbf{96.0} & 98.2 & \textbf{69.87} & 79.25 \\
    ECoT-trained, with generation & 95.6 & \textbf{98.3} & 68.72 & 79.12 \\
    \bottomrule
    \end{tabular}
\end{table}

Without ECoT generation at inference, ECoT training raises Long-suite success from 92.8\% to 96.0\% on LIBERO and from 65.18\% to 69.87\% on LIBERO-Plus, while LIBERO average success increases from 97.1\% to 98.2\%. Generating ECoT at inference provides no consistent additional gain: LIBERO average success increases slightly to 98.3\%, but Long-suite success decreases to 95.6\% and 68.72\%, respectively. Aggregate LIBERO-Plus success is 79.25\% without generation and 79.12\% with it, both below the baseline's 79.59\%.

\paragraph{Exploratory Object-Pose Supervision.}
Object-centric predictive methods use anticipated object configurations to guide manipulation: Goal-VLA recovers target object poses from generated goal images~\citep{chen2025goalvla}, while P2P-T learns future tool-pose representations to condition action generation~\citep{wang2026p2pt}. Motivated by these approaches, we examine whether future object-pose prediction as an auxiliary training objective improves action success on LIBERO-Plus. Two auxiliary heads share a pooled representation of VLM-token hidden states: a classification head predicts whether the object moves, and a regression head predicts its translation and rotation increments over the next eight action steps. Both heads are used only during training. Table~\ref{tab:object_pose_ablation} compares this variant with the SFT baseline; neither uses subtask conditioning at inference.

\begin{table}[H]
    \centering
    \reporttablestyle
    \setlength{\belowcaptionskip}{5pt}
    \caption{Auxiliary object-pose supervision on LIBERO-Plus by perturbation category (success rate, \%).}
    \label{tab:object_pose_ablation}
    \setlength{\tabcolsep}{2.5pt}
    \begin{tabular}{@{}lrrrrrrrr@{}}
    \toprule
    Variant & \makecell{Sensor\\noise} & \makecell{Camera\\viewpoints} & \makecell{Robot\\initial states} & \makecell{Light\\conditions} & \makecell{Language\\instructions} & \makecell{Background\\textures} & \makecell{Object\\layout} & Overall \\
    \midrule
    SFT baseline & 95.57 & 88.18 & \textbf{75.16} & \textbf{99.12} & \textbf{90.70} & \textbf{99.44} & 86.49 & 89.93 \\
    \makecell[l]{+ Object-pose\\supervision} & \textbf{96.56} & \textbf{89.56} & 74.52 & 98.25 & 90.50 & 98.88 & \textbf{87.15} & \textbf{90.12} \\
    \bottomrule
    \end{tabular}
\end{table}

Object-pose supervision increases overall success from 89.93\% to 90.12\%, a gain of 0.19 percentage points. Success improves under camera-viewpoint, sensor-noise, and object-layout perturbations, with the largest gain of 1.38 percentage points under viewpoint changes. It decreases in the other four categories, including drops of 0.87 percentage points under lighting changes and 0.64 under robot-initial-state changes.
\endgroup

\section{Conclusion}
We presented \ourmodel{}, a VLA model that uses future prediction to learn compact physical representations for action generation. Direct supervision and a pretrained video world model shape these representations during training; at deployment, the policy uses the learned physical tokens without running the video model or prediction heads. Results across four simulation benchmarks demonstrate strong manipulation performance, while the prediction study shows that world-model assistance improves accuracy over direct supervision alone.

Future work will explore predictive targets beyond RGB, depth, and optical flow, including object-centric geometry, contact states, and material properties. We will investigate how these cues complement one another and contribute to action learning while retaining a lightweight policy.

\clearpage
\bibliographystyle{unsrtnat}
\bibliography{references}  

\clearpage
\appendix
\section*{Appendix}

\section{Implementation and Training Details}
\label{app:training_details}

\subsection{Action-Expert Configuration}
The action expert has 16 Transformer blocks, alternating cross-attention and self-attention, with a token width of 768 and 12 attention heads of width 64. A final token projection maps features to 1024 dimensions before they are decoded into action velocities. The expert's input sequence consists of a proprioceptive prefix token, 32 learnable tokens, and 8 action tokens with learned positional embeddings. Attention and feed-forward dropout is 0.2, and the proprioceptive input is dropped with probability 0.2 during training.

During action flow-matching training, we randomly sample the flow time $\tau$ to vary the mixture of Gaussian noise and target actions in Eq.~\ref{eq:action_flow_path}. Smaller values give more weight to noise, while larger values give more weight to the target actions. Specifically, we draw $b\sim\operatorname{Beta}(1.5,1)$ and set $\tau=1-\min(b,s)/s$, with $s=0.999$. Flow time is discretized into 1000 buckets and embedded both in the action encoder and in the adaptive normalization of the Transformer blocks and final output projection. Each example is repeated 8 times with independently sampled flow times and noise for the flow-matching objective. Inference uses four Euler steps of size 0.25 over $[0,1]$.

\subsection{Frame Selection}
We select starting frames for VLA training at regular intervals of approximately 0.125~s (8~Hz). We also add frames where a gripper or dexterous-hand target command changes by more than a threshold between consecutive frames. Each manipulator's available gripper and hand channels are checked separately. We combine the two sets of candidates and remove frames that are too close together, keeping frames with command changes in preference to regularly sampled frames. The minimum spacing is 0.01~s, converted to a frame interval using the episode's frame rate. The remaining candidates must pass the sample-validity checks described in Section~\ref{subsec:policy_adaptation}.

\subsection{World-Model-Assisted Pretraining Comparison}

\paragraph{Training Setup.}
The comparison in Section~\ref{sec:ablation_analysis} tests whether adding world-model supervision improves future prediction. Both runs use direct future-feature supervision; Direct + WM additionally includes the auxiliary world-model objective. The two runs share the following settings. Training clips contain 5 frames at a resolution of $480\times640$. We optimize the physical module and trainable prediction heads with AdamW, a peak learning rate of $3\times10^{-4}$, and weight decay 0.01. The learning rate increases linearly over the first 5\% of updates and then follows cosine decay. Both runs use 8 GPUs with a global batch size of 128 and random seed 42; trainable parameters remain in FP32 with BF16 autocasting. Each run consists of 35{,}082 updates and 4{,}490{,}400 sample exposures, with 32 examples in the final update. The direct-loss weight increases linearly from zero to 0.5 over the first 2{,}000 updates in both runs.

\paragraph{Evaluation Protocol.}
We evaluate the trained physical modules and their original direct prediction heads without refitting or updating parameters. Both runs use the same 828 RoboCOIN clips from 100 held-out episodes and identical target-normalization statistics. Checkpoints are evaluated at steps 0, 1{,}000, 2{,}000, 5{,}000, 10{,}000, 20{,}000, 30{,}000, and 35{,}082; Table~\ref{tab:physical_wm_endpoint} reports the final checkpoint. Modality errors are averaged over clips, and the overall mean gives equal weight to RGB, depth, and optical flow. Relative reductions use Direct-only error as the denominator. We estimate 95\% percentile intervals from 10{,}000 paired episode-bootstrap resamples using seed 42, retaining all clips of each sampled episode and computing clip-weighted means within each resample. These intervals quantify episode-sampling uncertainty for the fixed trained checkpoints.

\FloatBarrier
\section{Physical-Module Design Choices}
\label{app:physical_design}

We select four physical-module settings in sequence: the VLM input layer, cross-attention placement, direct-supervision weighting, and query count. At each stage, we retain the settings selected in the preceding stages and compare candidates for the next choice. For each candidate, we train the physical module with the original Qwen3.5-4B backbone, then freeze both the VLM and physical module and fit a new linear readout to predict future features. The readout measures how readily future information can be recovered from the learned representations. We use validation scores to select configurations, with the single-seed results summarized in Table~\ref{tab:physical_design_ablation}.

\paragraph{Training Setup and Comparison Protocol.}
The design study uses a 500k-example training pool, 6{,}000 optimization steps, seed 42, and learning rate $3\times10^{-4}$. Physical-module hidden width, block count, head count, output width, and feed-forward expansion ratio are fixed at 1024, 4, 4, 2560, and 1. The world-model-assisted loss has weight 1 throughout these design comparisons.

We first compare VLM input layers using 256 queries, cross-attention at every physical-module block, and a constant direct-loss weight of 0.1. We then fix the selected VLM Block 8 and compare cross-attention placements, retaining 256 queries and weight 0.1; parameter counts vary with the number of cross-attention blocks. Next, we retain Block 8, every-block cross-attention, and 256 queries while comparing direct-loss weights and warm-up schedules. Finally, we compare query counts using Block 8, every-block cross-attention, and the selected direct-loss schedule, which warms up to 0.5 over 2{,}000 steps.

\paragraph{Frozen-Representation Probe.}
During readout fitting and evaluation, the VLM and physical module remain frozen. Masked mean pooling produces a 2560-dimensional representation, from which an affine ridge readout predicts pooled future RGB, depth, and color-encoded flow latent targets. The probe data span five source families, with 2{,}048 clips for readout fitting and a separate set of 1{,}024 validation clips for scoring each configuration. Within the fitting pool, an episode-separated split of 1{,}665 fitting clips and 383 validation clips is used to select the regularization strength. We then refit the readout on all 2{,}048 fitting clips before scoring it on the separate validation set. Readout fitting uses FP64 standardized spectral ridge, with standardization statistics estimated from fitting data only.

We report Mean Prediction Skill, which measures the relative MSE reduction over a baseline that always predicts the fitting-set mean for each source family and modality. The reductions are averaged equally across source families and modalities. A score of zero matches this baseline, and higher scores indicate better prediction. Let $\mathcal{S}$ denote the source families, $\mathcal{M}=\{r,d,f\}$ the RGB, depth, and flow modalities, $\widehat y$ the readout prediction, and $y$ the target. For source $s$ and modality $m$, let $\bar y^{\mathrm{fit}}_{s,m}$ be the mean target estimated from fitting labels. The score is
\begin{equation}
    \frac{1}{|\mathcal{S}||\mathcal{M}|}
    \sum_{s\in\mathcal{S}}\sum_{m\in\mathcal{M}}
    \left[
    1-\frac{\operatorname{MSE}_{s,m}(\widehat y,y)}
    {\operatorname{MSE}_{s,m}(\bar y^{\mathrm{fit}}_{s,m},y)}
    \right].
    \label{eq:physical_probe_skill}
\end{equation}

\begin{table}[!htbp]
    \centering
    \reporttablestyle

    \caption{Physical-module design comparisons using frozen-representation probes.}
    \label{tab:physical_design_ablation}
    \begin{tabular}{@{}lr@{}}
    \toprule
    Design choice & Mean Prediction Skill $\uparrow$ \\
    \midrule
    \multicolumn{2}{@{}l}{\textit{VLM input layer}} \\
    Block 8 & \textbf{0.3617} \\
    Block 16 & 0.3413 \\
    Block 24 & 0.3314 \\
    Final block & 0.2432 \\
    \addlinespace
    \multicolumn{2}{@{}l}{\textit{Cross-attention placement within the physical module}} \\
    First block & 0.3383 \\
    Final block & 0.3590 \\
    Every block & \textbf{0.3617} \\
    \addlinespace
    \multicolumn{2}{@{}l}{\textit{Direct prediction supervision}} \\
    No direct supervision & 0.2683 \\
    Constant weight 0.1 & 0.3617 \\
    Constant weight 0.5 & 0.3769 \\
    Constant weight 1.0 & 0.3723 \\
    Linear warm-up to 0.5 over 2{,}000 steps & \textbf{0.3834} \\
    \addlinespace
    \multicolumn{2}{@{}l}{\textit{Number of learned queries}} \\
    50 & \textbf{0.3844} \\
    128 & 0.3815 \\
    256 & 0.3834 \\
    \bottomrule
    \end{tabular}
\end{table}

\paragraph{Results and Interpretation.}
Block 8 gives the highest score among the tested VLM layers (0.3617 versus 0.2432 for the final block), supporting intermediate-layer input. Cross-attention at every physical-module block scores slightly higher than final-block-only access (0.3617 versus 0.3590). With world-model supervision retained, adding direct supervision with the selected weight warm-up raises the score from 0.2683 to 0.3834, the highest among the tested weighting schemes. Under this schedule, 50 queries achieve 0.3844, close to 0.3834 for 256 queries, supporting the compact token sequence.

\clearpage
\section{Additional LIBERO-Plus Results}
\label{app:auxiliary_supervision_details}

We additionally summarize the LIBERO-Plus results by the originating LIBERO task suite (Table~\ref{tab:subtask_pose_suites}). These are the same evaluations reported by perturbation category in Tables~\ref{tab:subtask_ablation} and~\ref{tab:object_pose_ablation}, grouped to show how the effects vary across task suites.

\begin{table}[H]
    \centering
    \reporttablestyle

    \caption{Subtask conditioning and object-pose supervision on LIBERO-Plus by source task suite (success rate, \%).}
    \label{tab:subtask_pose_suites}
    \begin{tabular}{@{}lrrr@{}}
    \toprule
    Source task suite & SFT baseline & \makecell{+ Subtask\\conditioning} & \makecell{+ Object-pose\\supervision} \\
    \midrule
    Spatial & 96.50 & \textbf{96.59} & 95.71 \\
    Object  & 92.85 & \textbf{93.21} & 93.05 \\
    Goal    & 83.94 & 84.06 & \textbf{85.18} \\
    Long    & 86.90 & \textbf{87.53} & 86.94 \\
    \midrule
    Overall               & 89.93 & \textbf{90.23} & 90.12 \\
    \bottomrule
    \end{tabular}
\end{table}

Subtask conditioning improves success across all four source suites, with its largest gain on Long ($+0.63$ percentage points). Object-pose supervision gives its largest gain on Goal ($+1.24$ percentage points) but reduces Spatial success by 0.79 percentage points.


\section{Franka End-Effector Pose Distribution Diagnostics}
\label{app:pose_distribution_audit}

We examine absolute end-effector (EEF) height distributions in the Franka data from the pretraining mixture and LIBERO, and assess whether the normalization range from pretraining covers the LIBERO observations. The comparison focuses on distribution differences associated with scene geometry, even when the robot embodiment is unchanged.

\paragraph{Audit Protocol.}
We inspect 75 Franka configuration entries from the pretraining mixture and the four LIBERO suites. For each entry, we sample up to eight shards and eight row groups per selected shard, targeting approximately 2{,}048 frames with seed 42. The audit yields 143{,}507 valid arm-level poses from 162{,}081 sampled frames; seven RobotSet entries lack EEF poses and are recorded as missing. Positions are interpreted in meters according to source metadata and retained in their original coordinate frames.

\paragraph{Scene Geometry and Normalization Ranges.}
The pretraining right-arm position statistics examined in the audit have a first-to-99th-percentile $z$ interval of approximately $[-0.096,0.772]$~m. We use its upper bound of 0.772~m as a reference to examine the coverage of LIBERO EEF heights.

LIBERO Spatial and Goal use tabletops at 0.90~m, Object uses a floor-level workspace, and Long combines living-room, kitchen, and study scenes with work surfaces at 0.41, 0.90, and 0.867~m, respectively. Checks of the original NVIDIA LIBERO position fields confirm that the suite-level height differences are present before pose conversion. The suite-dependent height ranges are consistent with these scene geometries. Table~\ref{tab:libero_pose_distribution} compares the sampled heights with the examined pretraining upper bound.

\begin{table}[H]
    \centering
    \reporttablestyle
    \caption{Sampled absolute EEF-height statistics in LIBERO. The pretraining upper bound is 0.772~m.}
    \label{tab:libero_pose_distribution}
    \begin{tabular}{@{}lrrr@{}}
    \toprule
    Suite & Median $z$ (m) & $z$ p01--p99 (m) & Above upper bound (\%) \\
    \midrule
    Spatial & 1.041 & 0.915--1.285 & 100.0 \\
    Object  & 0.225 & 0.010--0.336 & 0.0 \\
    Goal    & 1.076 & 0.911--1.290 & 100.0 \\
    Long    & 0.712 & 0.447--1.267 & 49.1 \\
    \bottomrule
    \end{tabular}
\end{table}

All sampled Spatial and Goal positions and 49.1\% of sampled Long positions exceed the reference upper bound. The differences across suites illustrate how a shared robot embodiment can still have different absolute pose distributions because of scene geometry, affecting the coverage of a normalization range transferred between datasets.

\end{document}